\documentclass{article}
\usepackage{ijcai24}

\usepackage{times}
\usepackage{soul}
\usepackage{url}
\usepackage[hidelinks]{hyperref}
\usepackage[utf8]{inputenc}
\usepackage[small]{caption}
\usepackage{graphicx}
\usepackage{amsmath}
\usepackage{amsthm}
\usepackage{booktabs}
\usepackage{algorithm}
\usepackage{algorithmic}
\usepackage[switch]{lineno}

\usepackage{enumitem}
\usepackage{dsfont}
\usepackage{pdfpages}
\usepackage{booktabs} 
\usepackage{siunitx} 
\usepackage{multirow} 
\usepackage{amssymb} 
\usepackage{caption} 
\usepackage{xspace}
\usepackage[]{subcaption} 
\usepackage{tabularx} 
\newcolumntype{R}{>{\raggedleft\arraybackslash}X}
\usepackage{bm} 
\usepackage{makecell}
\usepackage[export]{adjustbox}
\usepackage{newfloat}
\usepackage{makecell}

\newcommand{\ie}{\emph{i.e.}}

\makeatletter
\DeclareRobustCommand\onedot{\futurelet\@let@token\@onedot}
\def\@onedot{\ifx\@let@token.\else.\null\fi\xspace}

\def\ie{\emph{i.e\onedot}} 

\makeatother

\title{Benchmarking Fish Dataset and Evaluation Metric in Keypoint Detection - Towards Precise Fish Morphological Assessment in Aquaculture Breeding}
\author{
    Author Name
    \affiliations
    Affiliation
    \emails
    email@example.com
}

\iftrue
\author{
    Weizhen Liu$^1$ \and
    Jiayu Tan$^1$ \and
    Guangyu Lan$^1$ \and
    Ao Li$^1$ \and
    Dongye Li$^4$ \and
    Le Zhao$^1$ \and\\
    Xiaohui Yuan$^{1,3}$\thanks{Corresponding authors.} \And
    Nanqing Dong$^{2*}$
    \\
    \affiliations
    $^1$School of Computer Science and Artificial Intelligence, Wuhan University of Technology\\
    $^2$Shanghai Artificial Intelligence Laboratory\\
    $^3$Yazhouwan National Laboratory\\
    $^4$Sanya Boruiyuan Technology Co. Ltd
    \\
    \emails
    \{liuweizhen, tjy2023305211,yuanxiaohui\}@whut.edu.cn,
    dongnanqing@pjlab.org.cn
}
\fi

\begin{document}

\maketitle

\begin{abstract}
    Accurate phenotypic analysis in aquaculture breeding necessitates the quantification of subtle morphological phenotypes. Existing datasets suffer from limitations such as small scale, limited species coverage, and inadequate annotation of keypoints for measuring refined and complex morphological phenotypes of fish body parts. To address this gap, we introduce FishPhenoKey, a comprehensive dataset comprising 23,331 high-resolution images spanning six fish species. Notably, FishPhenoKey includes 22 phenotype-oriented annotations, enabling the capture of intricate morphological phenotypes. Motivated by the nuanced evaluation of these subtle morphologies, we also propose a new evaluation metric, Percentage of Measured Phenotype (PMP). It is designed to assess the accuracy of individual keypoint positions and is highly sensitive to the phenotypes measured using the corresponding keypoints. To enhance keypoint detection accuracy, we further propose a novel loss, Anatomically-Calibrated Regularization (ACR), that can be integrated into keypoint detection models, leveraging biological insights to refine keypoint localization. 
    Our contributions set a new benchmark in fish phenotype analysis, addressing the challenges of precise morphological quantification and opening new avenues for research in sustainable aquaculture and genetic studies. Our dataset and code are available at https://github.com/WeizhenLiuBioinform/Fish-Phenotype-Detect.
\end{abstract}

\section{Introduction}
\label{sec:intro}
Fish plays a significant role in the human diet due to their high-quality protein and rich nutrients. Over the past few decades, aquaculture has emerged as the fastest-growing sector in global agriculture, with fish becoming of utmost importance~\cite{FAO2022}. The availability of fish provides diverse and balanced food choices, which is crucial in addressing hunger and poverty. Measuring morphological phenotypes is indispensable in fish farming and breeding, as they provide valuable information for population studies, functional gene research, fish health assessment, the development of new strains, and the implementation of sustainable aquaculture practices~\cite{verhaegen2007deformities,figueroa2018novel,castrillo2021morphopathology}. The study of these measurements aligns with the United Nations' Sustainable Development Goals (SDGs), including No Poverty, Zero Hunger, and Life Below Water~\cite{UN2023sdg}.

\begin{figure}[t]
 \centering
 \subfloat[Grouper]{
   \includegraphics[width=0.19\textwidth]{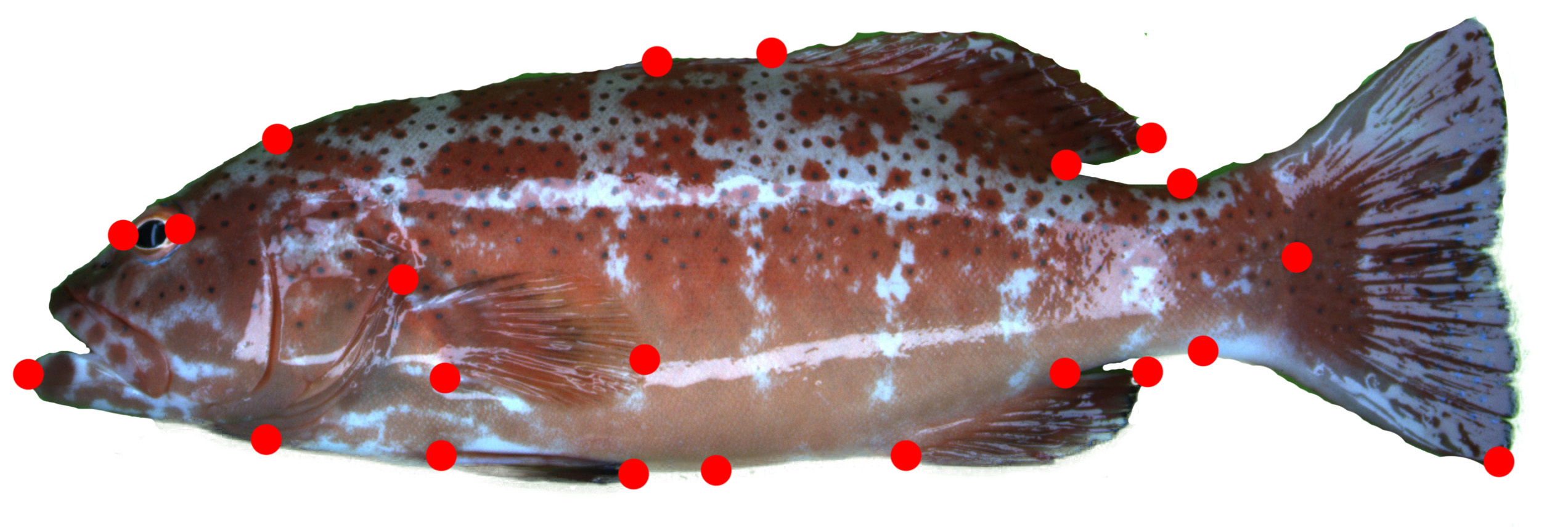}
 }
 \hfill
 \subfloat[Mottled naked carp]{
  \includegraphics[width=0.19\textwidth]{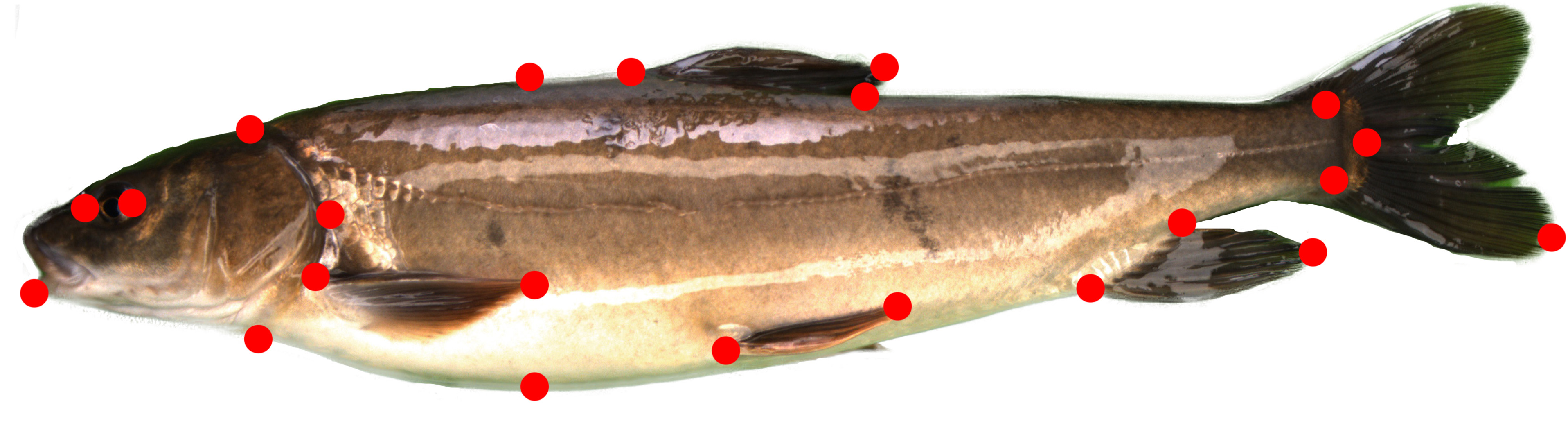}
 }
\\
 \subfloat[Bighead carp]{
 \includegraphics[width=0.19\textwidth]{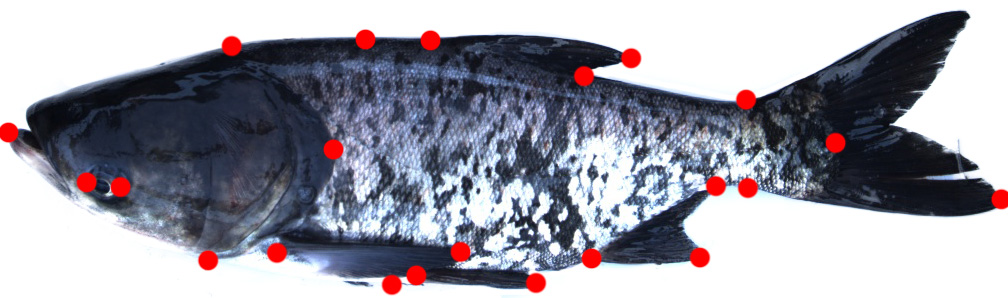}
}
  \hfill
  \subfloat[Common carp]{
  \includegraphics[width=0.19\textwidth]{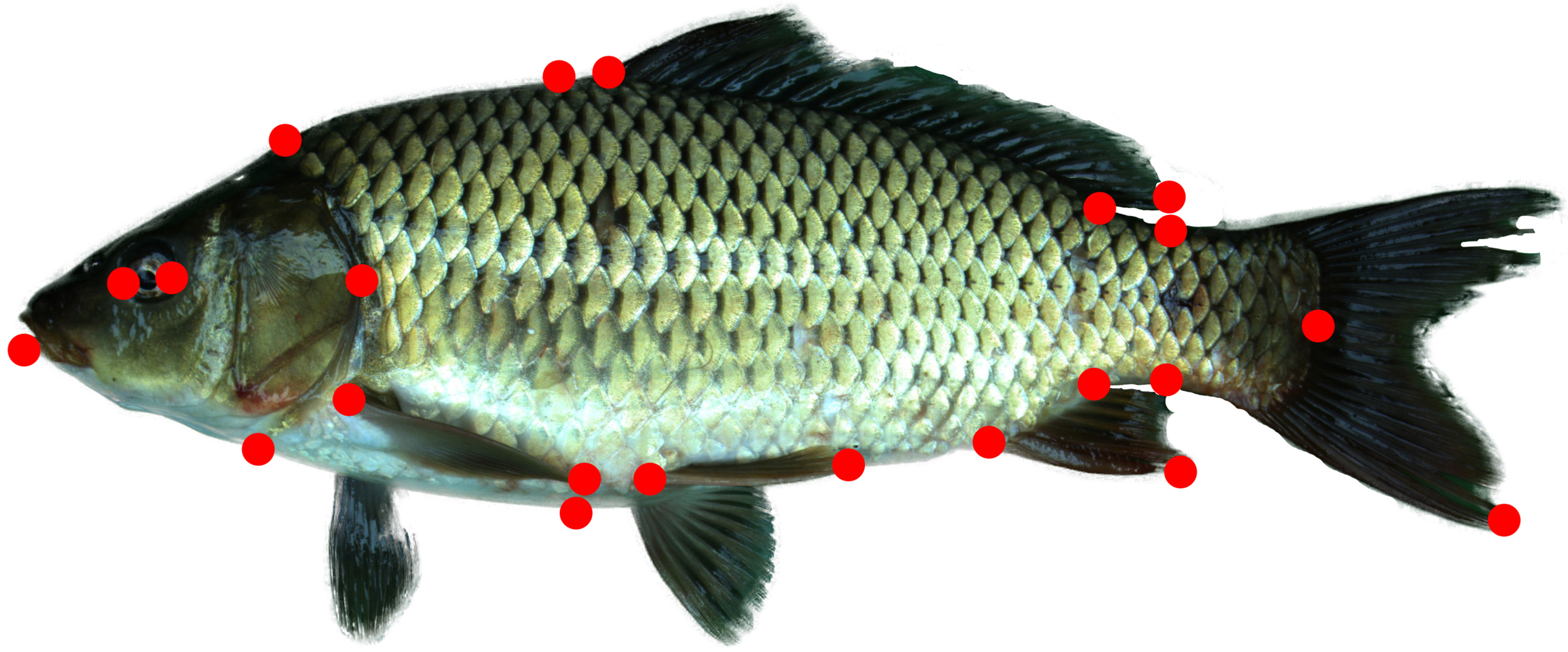}
 }
\caption{Examples of four fish species with 22 defined keypoints. Each red point represents an anatomical keypoint.} 
\label{fig:fish_keypoint}
\end{figure}

Detecting anatomical keypoints is a crucial step in conducting morphological assessment in various bioimage studies~\cite{pisov2020keypoints,devine2020registration,mitteroecker2022thirty}. Firstly, keypoints on fish bodies are important indicators for biological studies. Secondly, keypoints can be used to calculate specific morphological phenotypes. For instance, the head length, relevant to assessing the size and shape of the head, is measured as the distance from the snout tip to the posterior margin of the operculum. This phenotype provides information about the feeding behavior, sensory perception, and phylogenetic relationships of fish~\cite{Jerry1998MorphologicalVI,saleh2023mfld}. It is worth mentioning that increasing head length is a breeding goal for fish species like the bighead carp. 


Annotations of keypoint locations and measurements of morphological phenotypes in fish bioimage analyses have traditionally relied on manual processes. However, recent advancements in computer vision technology have prompted attempts to employ machine learning methods for these tasks~\cite{stern2011automatic,suo2020fish,kumar2022empirical,Dong2023detection,saleh2023mfld}. While current studies have successfully measured several morphological phenotypes related to the shape and size of fish body contours using detected keypoints, quantifying more refined and complex morphological phenotypes of specific body parts, such as eye diameter and caudal peduncle height, remains challenging. The main reason is the lack of annotated keypoints for measuring these subtle phenotypes. Moreover, existing datasets are relatively small in scale, and are often limited to a single fish species.

To overcome the aforementioned limitations, this paper presents the construction and release of a high-quality, large-scale, and multi-species fish keypoint detection dataset, referred to as \textbf{FishPhenoKey}. 
FishPhenoKey comprises 23,331 fish images from six common fish species: grouper, bighead  carp, common carp, mottled naked carp, longsnout catfish, and spotbanded scat. For the first four species, 10,327 images are annotated with 22 anatomical keypoints specifically defined for measuring up to 23 subtle morphological phenotypes of interest to fish breeders. Sample images with the annotated keypoints are shown in Fig.~\ref{fig:fish_keypoint}.

\begin{figure}[t]
 \centering
 \includegraphics[width=\columnwidth]{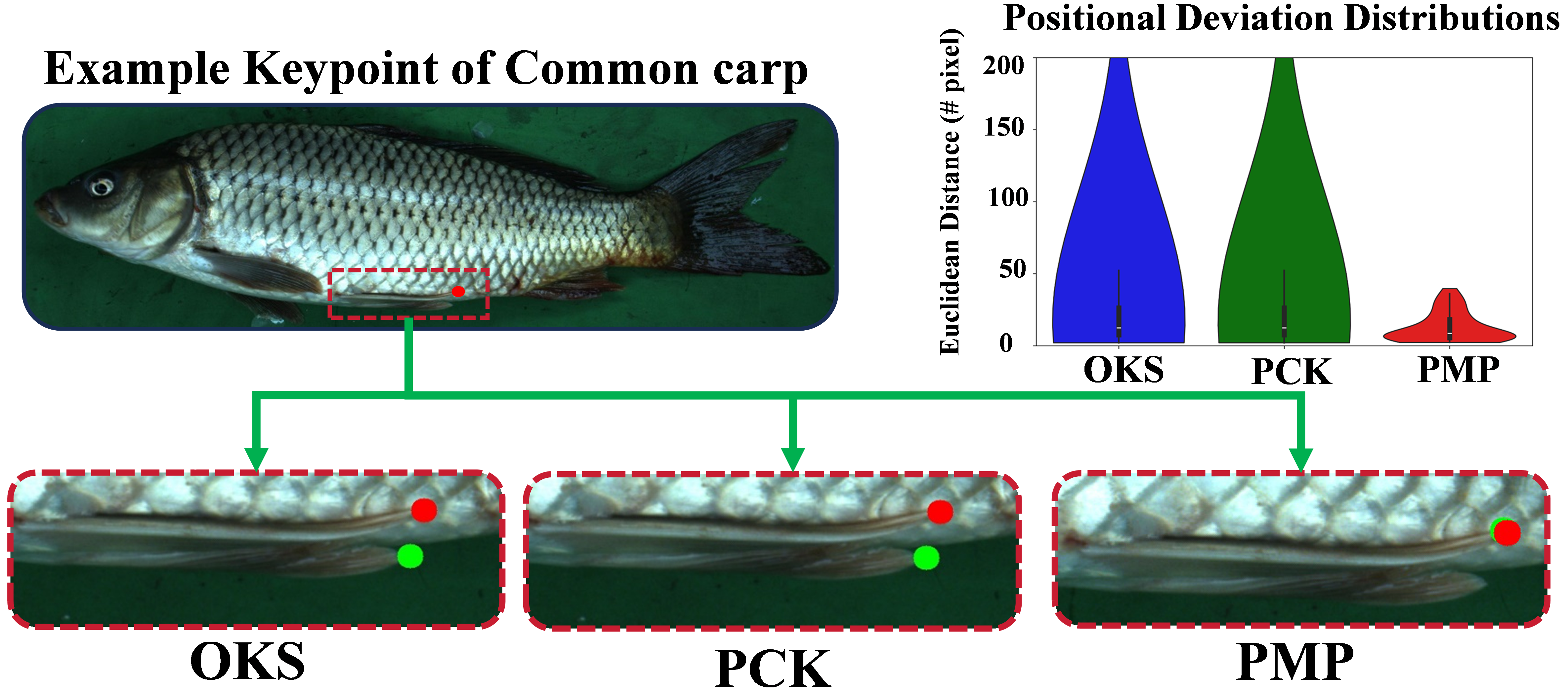}
 \caption{Visualization of the impact of evaluation metrics on keypoint prediction accuracy. The red dot represents the true position of the keypoint, while the green dot represents the predicted position. The violin charts show the distributions of positional deviations calculated as Euclidean distance (measured as the number of pixels) between predictions and ground truths of keypoints using three evaluation metrics. Notably, PMP demonstrates superiority due to its smallest positional deviations.}
 \label{fig:fish_metrics}
 \end{figure}

Furthermore, a dedicated evaluation criterion that is specifically designed to facilitate the nuanced evaluation of fish subtle morphologies is developed. The most popular evaluation criteria for fish keypoint detection now are the object keypoint similarity (OKS), which is widely utilized for human pose estimation to evaluate the overall performance of all keypoints~\cite{suo2020fish,saleh2023mfld,Dong2023detection,yu2023key}. However, our investigations show that, even with high OKS scores, there can be significant deviations in the measurements of morphological phenotypes using the detected keypoints. Therefore, a robust evaluation metric that takes into account the evaluation of individual keypoint position deviations is desired. Indeed, the task of interest requires extremely accurate localization of each keypoint, as even slight deviations in the keypoint positions can lead to significant deviations in the corresponding phenotype measurements. The existing evaluation metrics for assessing the performance of individual keypoints, such as Percentage of Correct Keypoints (PCK), are also originally developed for human pose estimation~\cite{andriluka20142d,Zhang_2019_CVPR,toshev2014deeppose}. However, these metrics are not fully appropriate for our fish keypoint detection task, as they lack sensitivity to the diverse phenotypes of different fish species. Therefore, we introduce a new phenotype-based metric, Percentage of Measured Phenotype (PMP). A qualitative comparison between the best model weights selected by different evaluation metrics is visualized in Fig.~\ref{fig:fish_metrics}, where the proposed phenotype-based metric does show more robust performance than general-purpose metrics.

To enhance the accuracy of keypoint detection towards fish morphological analysis, our research introduces an Anatomically-Calibrated Regularization (ACR) loss. Inspired by the anatomical prior in the literature of medical image analysis~\cite{dong2022towards},
ACR integrates fish-specific anatomical insights to achieve accurate localization. Specially, ACR introduces a novel component that accounts for the spatial relationships between different keypoints on fish body, thereby ensuring biologically precise keypoint localization. By incorporating Gradnorm~\cite{chen2018gradnorm}, ACR can balance the contribution of each loss component, aligning their gradients for harmonized learning. The empirical validation across diverse fish species not only demonstrates the effectiveness of ACR loss in improving phenotypic measurement precision but also highlights its adaptability and network-agnostic utility in addressing the complex nature of fish morphology.


We summarize our contributions as follows:
\begin{itemize}
\item We release the FishPhenoKey dataset, which includes 23,331 high-resolution images of six fish species, annotated with 22 phenotype-oriented keypoints to accurately capture subtle morphological phenotypes of fish body parts. 
\item We propose a new evaluation metric, PMP, which is sensitive to deviations in phenotype measurements, ensuring accurate assessments of the intricate phenotypic variations.
\item We introduce a new loss, ACR, incorporating biological insights to enhance the precision of keypoint detection and phenotype measurement across diverse fish species.
\item We extensively evaluate the proposed dataset and evaluation metric, establishing a benchmark for subsequent work and facilitating future research.
\end{itemize}

\section{Related Work}
\subsection{Fish Keypoint Datasets}
There are only a few publicly available fish keypoint datasets. To the best knowledge, these datasets do not include phenotype-oriented keypoints. Tab.~\ref{tab:dataset_descirption} summarizes the limitations of these datasets, focusing on aspects such as image clarity, species diversity, image modality, number of keypoints, and whether the detected keypoints are designed for measuring morphological phenotypes. For instance, the fish keypoints dataset~\cite{Dong2023detection} presents low-resolution underwater images, resulting in inadequate image clarity to depict distinct anatomical structures on fish bodies. Furthermore, the zebrafish microscopy dataset, medaka microscopy dataset, and seabream radiography dataset~\cite{kumar2022empirical} comprise keypoints annotated on the skeletal structure, which are used for classifying bone deformities rather than measuring morphological phenotypes.  FishPhenoKey is the first large-scale, multi-species, phenotype-oriented keypoint detection benchmark dataset.

\begin{table}[]
\centering
\setlength{\tabcolsep}{0.5pt}
\captionsetup{justification=justified,width=\columnwidth, labelfont=small, font=footnotesize}
\begin{tiny}
\begin{tabularx}{0.48\textwidth}{@{\extracolsep{\fill}}c>{\tiny}c>{\tiny}c>{\tiny}c>{\tiny}c>{\tiny}c>{\tiny}c@{\extracolsep{\fill}}}
\toprule
{\scriptsize {Dataset}} & {\scriptsize{\textbf{Pheno}}} & {\scriptsize {Clarity}} &  {\scriptsize{\# Species}}  & {\scriptsize {Modality}} & {\scriptsize{\# Keypoint}} & {\scriptsize {\# Img}}\\
\midrule
\makecell[c]{Fish keypoints dataset\\ \cite{Dong2023detection}} &  &  &  $-$ &  RGB &  7 &  2,000\\
\makecell[c]{Salmon Dataset\\ \cite{moccetti2023shape}} &  & \checkmark & 1  &  RGB &  22 &   \phantom{0}291\\ 
\makecell[c]{Zebrafish Microscopy Dataset\\ \cite{kumar2022empirical}} &  &  \checkmark &  1 &  Micro &  25 &   \phantom{0}113\\
\makecell[c]{Medaka Microscopy Dataset\\ \cite{kumar2022empirical}} &  &  \checkmark &  1 &  Micro &  6 &   \phantom{0}470\\
\makecell[c]{Seabream Radiography Dataset\\ \cite{kumar2022empirical}} &  &  \checkmark &  1 &  X-Ray &  19 &   \phantom{0}847\\
\midrule
FishPhenoKey (Ours) &  \checkmark &  \checkmark &  6 &  RGB &  22 &  \textbf{23,331}\\ 
\bottomrule
\end{tabularx}
\end{tiny}
\caption{Comparison of FishPhenoKey with existing fish keypoint datasets. ``Pheno'' means that the keypoints are annotated specifically for morphological phenotype measurements. ``Clarity'' refers to an image with sufficient clarity to display the structures of fish body parts. ``\#  Species'' denotes the number of fish species in the dataset. In the fish keypoints dataset, ``$-$'' is used because of the absence of fish species annotations, which prevents us from counting the number of species. ``Modality'' represents the image modality, ``Micro'' represents microscopy. ``\# Keypoint'' represents the number of different keypoints annotated.}
\label{tab:dataset_descirption}
\end{table}

\subsection{Keypoint Detection}
Keypoint detection has a long history in computer vision with various applications such as object recognition, image registration, and human pose estimation. Although deep learning-based methods such as HRNet~\cite{wang2020deep} and Hourglass~\cite{newell2016stacked} have achieved impressive accuracy in the human pose estimation field, their direct application to fish keypoint detection for precise measurement of all phenotypes remains a challenge. This is primarily due to the lack of constraints on the relative positions of fish keypoints during their training process. MFLD-net~\cite{saleh2023mfld} is currently the only proposed algorithm for fish phenotype-oriented keypoint detection. Unfortunately, it overlooks specific anatomical structures of fish, leading to significant room for improvement in accurately measuring fish phenotypes. Therefore, we developed a new loss function, ACR, specifically designed based on the anatomical structure of fish. The ACR loss aims to address the aforementioned limitations and improve the accuracy of fish phenotype measurement.

\subsection{Evaluation Metrics for Keypoint Detection}
Evaluating the performance of keypoint detection for morphological phenotype assessments can be challenging due to various criteria involved, including the number of keypoints, the number of keypoint-derived morphological phenotypes, the size of phenotypes, and the size of the fish body. OKS~\cite{lin2014microsoft} and PCK~\cite{yang2012articulated}, commonly used for evaluating the keypoint performances for human pose estimation, fail to deal with an important issue: the increasing difficulty of accurately measuring phenotype using keypoint detection as the size of the phenotype decreases. Additionally, OKS measures the average precision and recall for body-level detection~\cite{guesdon2021dripe}. To address these limitations, we define a novel phenotype-sensitive metric for evaluating individual keypoint called PMP.

\section{Preliminaries}
\label{sec:prelim}
This section describes two evaluation metrics mentioned in Sec.~\ref{sec:intro}, which are commonly used in keypoint detection.

\noindent \textbf{Object Keypoint Similarity} (OKS) measures the similarity between the predicted keypoints and the corresponding ground truths by calculating the distance between them, normalized by the scale of the object and a per-keypoint constant~\cite{lin2014microsoft}.
\begin{equation}
    \mathrm{OKS}=\frac{\sum_i \mathrm{KS}_i \cdot \sigma(v_i>0)}{\sum_i \sigma(v_i>0)}
\label{eq:oks}
\end{equation}
where $\sigma(v_i>0)$ indicates whether $i$-th keypoint is visible and annotated in the image, and $\text{KS}_i$ is the keypoint similarity, defined as:
\begin{equation}
\mathrm{KS}_i = \exp\left(-\frac{d_i^2}{2 \cdot s^2 \cdot k_i^2}\right),
\label{eq:ks}
\end{equation}
where $d_{i}$ is the Euclidean distance between the predicted and ground truth keypoints, $s$ is the scale of the object, and $k_i$ is a constant that standardizes the deviation for each body part. 

\noindent \textbf{Percentage of Correct Keypoints} (PCK) measures the accuracy of keypoint detection by calculating the percentage of keypoints that are correctly identified within a certain threshold distance from their ground truth positions~\cite{yang2012articulated}.
\begin{equation}
    d_{i,n}=\frac{\|p_{i,n}-g_{i,n}\|^2}{h_n}, 
\label{eq:pck}
\end{equation}
where $d_{i,n}$ denotes the Euclidean distance between the predicted keypoint $p_{i,n}$ and the ground truth keypoint $g_{i,n}$, normalized by the scale factor $h_n$, which is typically a particular body dimension or the overall scale of the object. Here, $n$ denotes the index of the sample in the dataset, and $N$ is the total number of samples.
\begin{equation}
\text{PCK}(i)=\sum_{n=1}^N\frac{\mathbb{I}(d_{i,n} < r)}{N}
\label{eq:pckh}
\end{equation}

$i$-th keypoint is correctly identified if $d_{i,n}$ is less than the pre-determined threshold $r$, which is often set as half the head's size (PCKh@0.5)~\cite{andriluka20142d} or a fifth of the torso's size (PCK@0.2)~\cite{johnson2010clustered}. $\mathbb{I}(d_{i,n} < r)$ is the indicator function controlling this process, which is formulated as:
\begin{equation}
\mathbb{I}(d_{i,n} < r) = 
\begin{cases} 
1 & \text{if } d_{i,n} < r \\
0 & \text{otherwise}
\end{cases}
\label{eq:indicator}
\end{equation}

\section{Benchmark Dataset}
\label{sec:dataset}
\noindent \textbf{Data Collection.}
We collect 23,331 high-quality fish images from six different fish species. The dataset consists of 17,511 images of grouper (\textit{Epinephelus}), 1,777 images of mottled naked carp (\textit{Gymnocypris eckloni}), 1,454 images of bighead  carp (\textit{Hypophthalmichthys nobilis}), 1,455 images of common carp (\textit{Cyprinus carpio}), 800 images of longsnout catfish (\textit{Leiocassis longirostris}), and 334 images of spotbanded scat (\textit{Selenotoca multifasciata}). These images were obtained by capturing live fish using our custom-designed image acquisition device in a controlled laboratory environment. The fish specimens are carefully positioned within the device, and their images are captured once they have acclimated to the environment and show no signs of stress. Our image acquisition device features a 360\textdegree\ LED white light source, a solid-color background, and high-performance CMOS industrial cameras, ensuring clear capture of the anatomical features of the fish body, facilitating precise annotation of keypoints and accurate measurement of morphological phenotypes. The images are saved in 24-bit RGB JPG format at resolutions of 4608 × 3456 and 1152 × 864 pixels.
\begin{figure}[t!]
    \centering
    \includegraphics[width=0.40\textwidth]{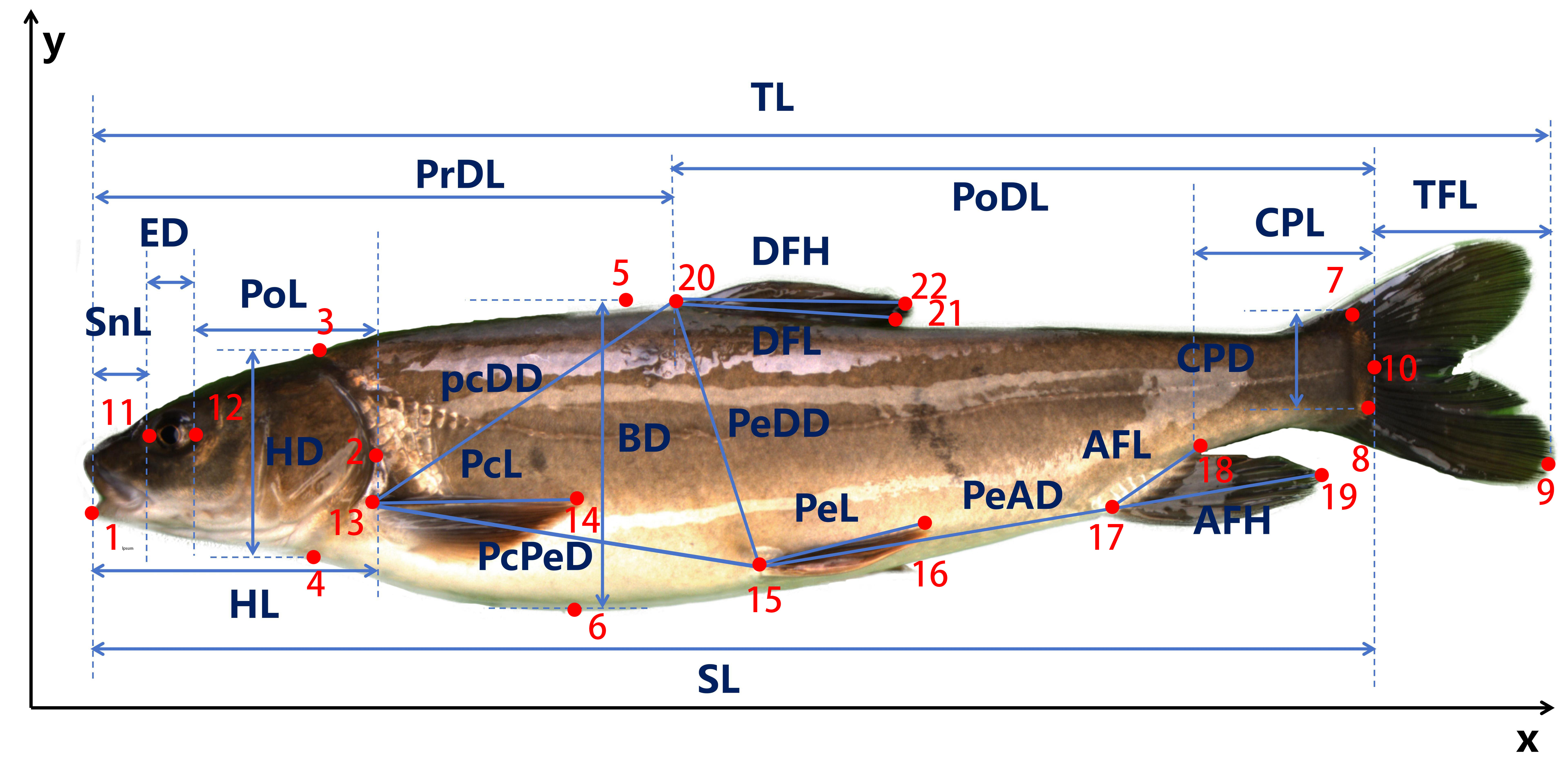}
    \caption{Examples of 22 annotated keypoints and their derived 23 morphological phenotypes on a mottled naked carp fish. Each red point represents a keypoint, and the red numbers 1-22 correspond to individual keypoint classes. Black abbreviations represent the phenotype names, totaling 23 phenotypes, and the blue solid lines indicate the relationships between the phenotypes and keypoints.}
    \label{fig:phenotype}
\end{figure}

\noindent \textbf{Annotations of Fish Keypoints.}
To measure subtle morphological phenotypes of fish body parts, 22 anatomical keypoints that are commonly present in multiple fish species are defined and manually annotated under the guidance of biological experts and descriptions in ~\cite{Jerry1998MorphologicalVI,liao20213dphenofish}. In the dataset, annotations were made for four out of six fish species, comprising a total of 10,327 images. Specifically, there are 9,112 images of grouper, 200 images of mottled naked carp, 706 images of bighead carp, 309 images of common carp. The definitions of 22 keypoints are as follows: 1: snout tip, 2: posterior end of operculum, 3: top end of head, 4: isthmus, 5: dorsal apex, 6: bottom end of ventral margin, 7: top end of caudal peduncle, 8: bottom end of caudal peduncle, 9: posterior end of tail fin, 10: posterior end of caudal vertebrae, 11: anterior end of eye, 12: posterior end of eye, 13: anterior end of pectoral fin, 14: posterior end of pectoral fin, 15: anterior end of pelvic fin, 16: posterior end of pelvic fin, 17: anterior end of anal fin, 18: posterior end of anal fin, 19: outer margin of anal fin, 20: anterior end of dorsal fin, 21: posterior end of dorsal fin, 22: outer margin of dorsal fin. Sample images with the annotated keypoints are shown in Fig.~\ref{fig:phenotype}.

\par
\par
\noindent \textbf{Fish Morphological Phenotypes.}
The 22 defined keypoints allow for measuring 23 morphological phenotypes. These phenotypes with the relationship of keypoints are shown in Fig.~\ref{fig:phenotype}, and definitions are as follows: TL: total length, SL: standard length, HL: head length, SnL: snout length, ED: eye diameter, PoL: postorbital length, BD: body depth, HD: head depth, PeAD: pelvic–anal fin origin distance, CPD: caudal peduncle depth, CPL: caudal peduncle length, DFL: dorsal fin length, DFH: dorsal fin height, PcL: pectoral fin length, PeL: pelvic fin length, AFL: anal fin length, AFH: anal fin height, TFL: tail fin length, PrDL: predorsal length, PoDL: postdorsal length, PcDD: pectoral–dorsal fin origin distance, PcPeD: pectoral–pelvic fin origin distance, PeDD: pelvic–dorsal fin origin distance.

\begin{figure*}[t]
    \centering
    \begin{subfigure}[b]{0.16\textwidth}
        \centering
        \includegraphics[height=1.4in]{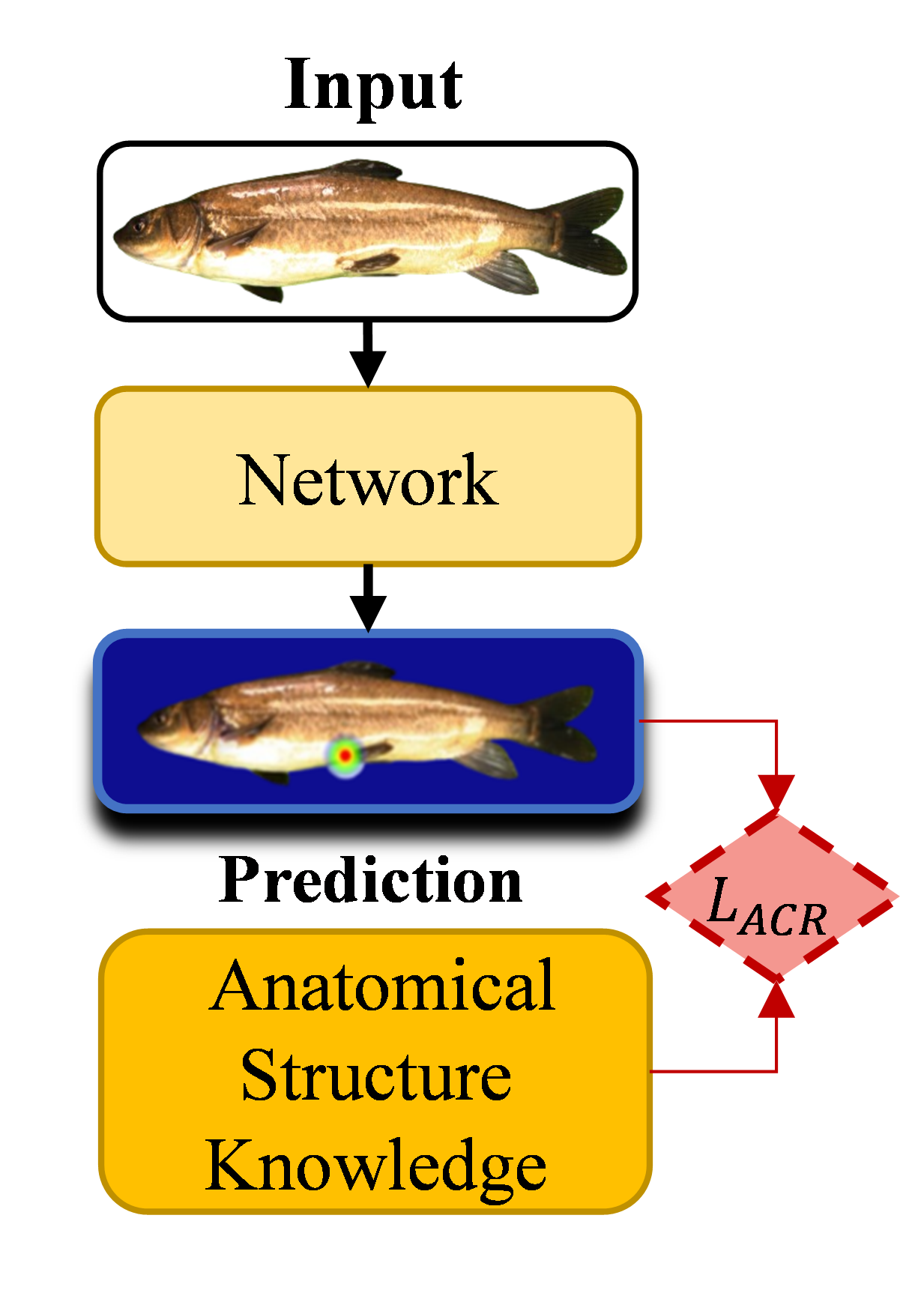}
        \caption{Framework}
        \label{fig:sub1}
    \end{subfigure}
    \begin{subfigure}[b]{0.62\textwidth}
        \centering
        \includegraphics[height=1.4in]{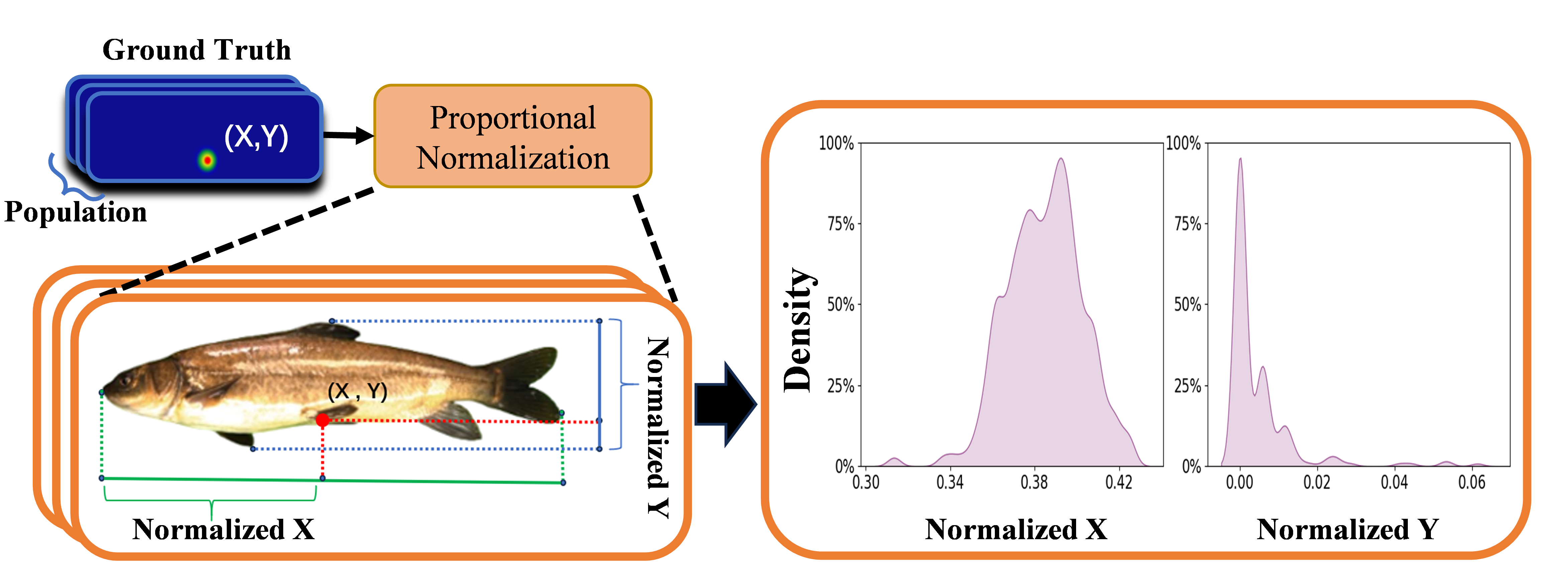}
        \caption{Anatomical Structure Knowledge}
        \label{fig:sub2}
    \end{subfigure}
    \begin{subfigure}[b]{0.21\textwidth}
        \centering
        \includegraphics[height=1.4in]{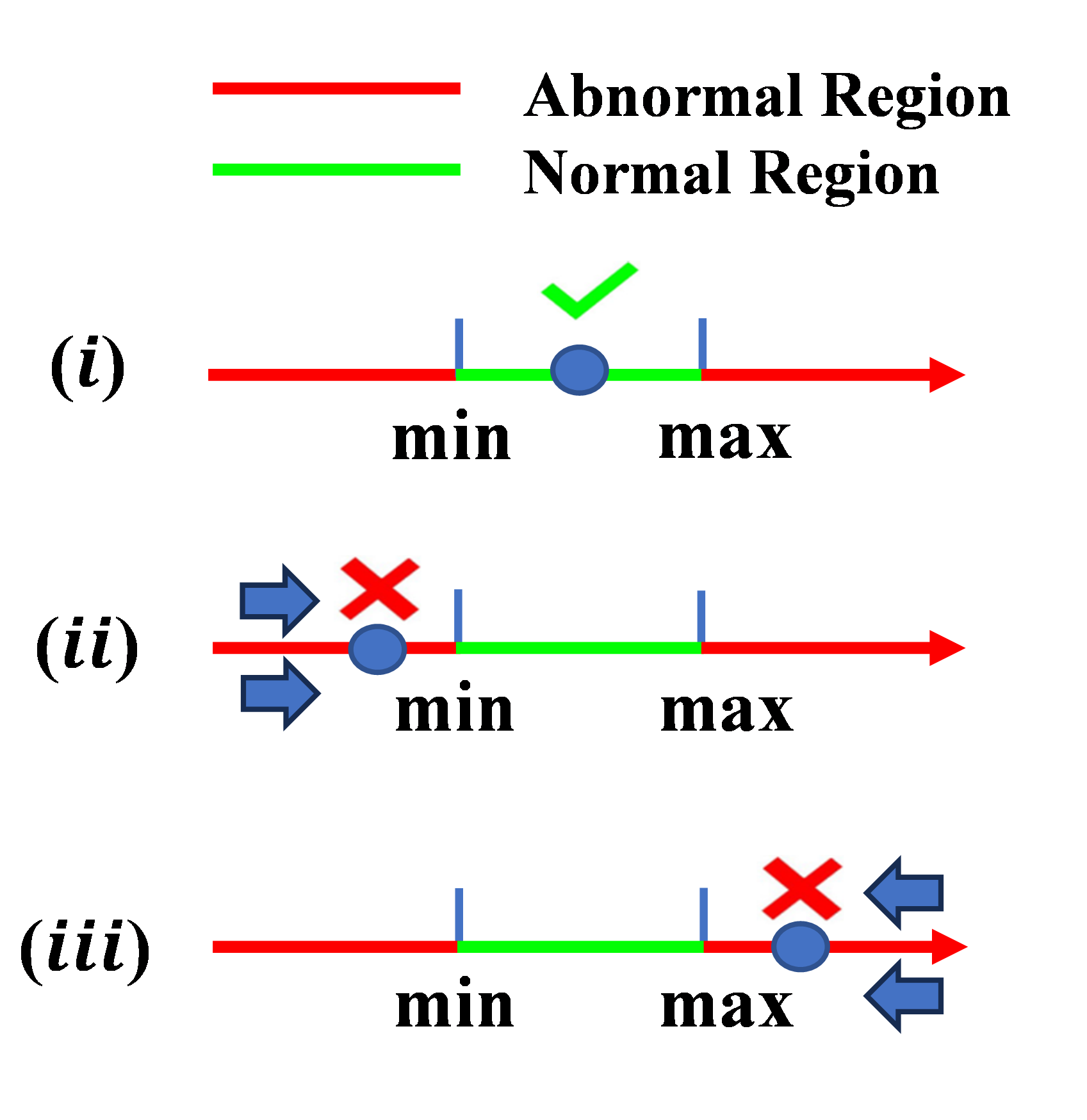}
        \caption{Motivation of $L_{ACR}$}
        \label{fig:sub3}
    \end{subfigure}
    \caption{(a) The overall framework for fish keypoint detection integrated with the proposed Anatomically-Calibrated Regularization (ACR) loss. (b) Extraction of fish anatomical structure knowledge from the ground truth keypoints for a given fish species. The distributions are the normalized coordinates of keypoint positions. The normalization process is described in Sec.~\ref{sec:method:acr}. (c) Conceptualized motivation for ACR loss. The green line segment denotes the ``normal'' region based on population study and the red one denotes the ``abnormal'' region. By minimizing the ACR loss, we aim to push the predicted keypoints from  ``abnormal'' region to ``normal'' region. 
    }
    \label{fig:main_fig}
\end{figure*}
\section{Method}
Our study introduces an innovative fish dataset, aiming to establish a novel benchmark in the field of computational ichthyology. We develop a custom evaluation metric specifically designed for keypoint detection with a focus on phenotype measurements. In this endeavor, we employ a keypoint detection network architecture to process images and annotations in COCO format~\cite{lin2014microsoft}, incorporating optimizations based on the biological features of fish to enhance keypoint recognition. This approach significantly improves the precision of keypoint localization, laying a solid foundation for detailed phenotypic analysis in later stages.

\subsection{Percentage of Measured Phenotype}


Currently, the primary application of keypoint detection tasks is in pose estimation. As a result, mainstream evaluation metrics such as PCK and OKS are specifically designed to assess the precision of each point in its measurement relative to the overall object. This is the reason behind the inclusion of a scaling factor in the denominator of Eq.~\eqref{eq:ks}, and incorporation of $h_{n}$ in the denominator of Eq.~\eqref{eq:pck}. However, these methods do not establish connections between keypoints. For phenotypic measurement tasks, the focus shifts to calculating the distance between keypoints, rendering the metrics based on object-scale unsuitable. So we introduce the Percentage of Measured Phenotype (PMP) metric. In contrast to Eq.~\eqref{eq:pck}, Eq.~\eqref{eq:Normalized Keypoint Distance} utilizes $pheno_{j,n}$, which represents the shortest related phenotype of the $j$-th keypoint instead of a fixed value $h_n$. This design indicates the different phenotypic characteristics associated with each keypoint.

\begin{equation}
    d_{j,n}=\frac{\left\|p_{j,n}-g_{j,n}\right\|^2}{pheno_{j,n}}
\label{eq:Normalized Keypoint Distance}
\end{equation}
PMP score for each keypoint, shown in Eq.~\eqref{eq:PMP}, is then calculated by averaging the binary results of $\mathbb{I} (d_{j,n} < r)$, where $\mathbb{I}$ is defined according to Eq.~\eqref{eq:indicator}. The threshold $r$ is set considering the biological variability and the expected precision for phenotypic measurements. The threshold $r$ is set at 0.1 in our experiments, reflecting the required precision for phenotypic measurements.Here, \(n\) denotes the index of the sample in the dataset, and \(N\) is the total number of samples, as previously mentioned.


\begin{equation}
   \mathrm{PMP}(j)=\sum_{n=1}^N\frac{\mathbb{I}(d_{j,n} < r)}{N}
 \label{eq:PMP}
 \end{equation}

The PMP metric redefines the approach to keypoint evaluation for tasks requiring phenotypic measurement by emphasizing biological relevance and the distances between keypoints.
\subsection{Anatomically-Calibrated Regularization Loss}
\label{sec:method:acr}
The relative positions of keypoints shall be consistent across species due to their biological roles, such as locomotion, feeding, and reproduction. This prior knowledge can be utilized to guide the model to localize keypoints in line with actual anatomical structures. In this study, we propose the Anatomically-Calibrated Regularization (ACR) loss, a module refining the traditional Mean Squared Error (MSE) loss by integrating anatomical structure knowledge as constraints in the training. By computing the normalized coordinates of the ground truth keypoints from the training data, we can acquire distributions that statistically describe the anatomical structure, shown in Fig.~\ref{fig:main_fig}(b). The ACR loss is designed to regularizing the model training to predict keypoints within a ``possible'' region defined by the statistical information of anatomical structure. The motivation is illustrated in Fig.~\ref{fig:main_fig}(c).

Let $(x_{i,j}, y_{i,j})$ denote the coordinate of the $i$-th keypoint in the $j$-th image, and $(x_{i,j}', y_{i,j}')$ is the normalized coordinate based on the fish body dimension, calibrated as follows: 
\begin{equation}
(x_{i,j}', y_{i,j}') = \left( \frac{x_{i,j} - x_{\min,j}}{x_{\max,j} - x_{\min,j}}, \frac{y_{i,j} - y_{\min,j}}{y_{\max,j} - y_{\min,j}} \right)
\end{equation}
where $x_{\min,j}$, $y_{\min,j}$, $x_{\max,j}$, and $y_{\max,j}$ refer to the minimum and maximum values of coordinates across all keypoints in the $j$-th image. To determine the size constraints of the $i$-th keypoint in the $j$-th image, we first ascertain its extremes, \ie~$\min_{\forall i,j}\{x_{i,j}'\}$, $\max_{\forall i,j}\{x_{i,j}'\}$, $\min_{\forall i,j}\{y_{i,j}'\}$, and $\max_{\forall i,j}\{y_{i,j}'\}$. Then, we multiply the extremes of these distributions by the fish's dimensions which are defined as the ranges, \ie~$x_{\max,j}-x_{\min,j}$ and $y_{\max,j}-y_{\min,j}$. Before defining the constraint, we introduce a hypothetical rectangular (box), where the coordinates of two vertices in the diagonal are defined as below.

\begin{equation}
\begin{split}
k_{i,j}^{min} = \bigg(& \min_{\forall i,j}\left\{x_{i,j}'\right\} \cdot \left(x_{\max,j} - x_{\min,j}\right), \\
& \min_{\forall i,j}\left\{y_{i,j}'\right\} \cdot \left(y_{\max,j} - y_{\min,j}\right)\bigg)
\end{split}
\label{eq:k_min}
\end{equation}
\begin{equation}
\begin{split}
k_{i,j}^{max} = \bigg(& \max_{\forall i,j}\left\{(x_{i,j}')\right\} \cdot \left(x_{\max,j} - x_{\min,j}\right), \\
& \max_{\forall i,j}\left\{(y_{i,j}')\right\} \cdot \left(y_{\max,j} - y_{\min,j}\right)\bigg)
\end{split}
\label{eq:kij_max}
\end{equation}
Intuitively, the minimum and maximum expected proportional positions of the $i$-th keypoint are equivalent to generating a box constraint, where $k_{i,j}^{min}$ is equivalent to the upper left corner coordinate of the box and $k_{i,j}^{max}$ is the lower right corner coordinate of the box. Utilizing these expected proportional positions, we further define the ACR loss for precise keypoint localization as follows:

\begin{equation}
\begin{split}
L_{ACR} = \sum_{i=1}^{22} (& \max(0, k_{i,j}^{min} - pred_{i,j}) \\
& + \max(0, pred_{i,j} - k_{i,j}^{max}) )
\end{split}
\end{equation}

These correspond to the range of actual positions recovered from the smallest and largest calibration values in the $k$-th  distribution, and $pred_{i,j}$ is the predicted position of the $i$-th keypoint in the $j$-th image.


The final optimization goal is a combination of $L_{MSE}$ and $L_{ACR}$. We simply use GradNorm~\cite{chen2018gradnorm} to automatically balance the training process.





\section{Experiments}
The main objectives of the experimental design on FishPhenoKey dataset are threefold. First, we aim to evaluate the robustness of PMP metric, which is designed to evaluate individual keypoints. Second, we aim to evaluate the effectiveness of ACR module for keypoint detection. Third, we aim to evaluate the effectiveness of ACR module for morphological phenotype measurement. By achieving these goals, we hope to establish a benchmark for future work and promote further research in this field. 

\subsection{Experimental Setup} 
All experiments are conducted using PyTorch~\cite{paszke2019pytorch} on NVIDIA 4090 GPU with 24GB of memory. HRNet~\cite{wang2020deep} serves as the backbone network architecture. 
The generation of the training input image is carried out in two steps: (1) Cropping and resizing the body regions of the fish
; (2) performing augmentations with random rotation (\(45^\circ\)) and random scaling (\(\pm 35\%\)). We employ the AdamW~\cite{kingma2015adam} optimizer with a batch size of 2, where $\boldsymbol{\beta}$s are set to 0.5 and a learning rate decay strategy using ReduceLROnPlateau\footnote{\url{https://pytorch.org/docs/stable/optim.html}}. To ensure a fair comparison, all experiments are initiated with a learning rate of $10^{-4}$ and a weight decay of $10^{-5}$. Each species is trained for 100 epochs using a dataset of 200 images. In this study, ``K-$\#$'' denotes the $k$-th keypoint. For example, ``K-6'' denotes the bottom end of ventral margin (see Sec.~\ref{sec:dataset} for details).
\begin{table}[t]
\centering
\captionsetup{font=footnotesize, width=\columnwidth, justification=justified}
\setlength{\tabcolsep}{2pt} 
\renewcommand{\arraystretch}{1.0}

\begin{tabularx}{\columnwidth}{@{}l *{4}{>{\centering\arraybackslash}X}@{}} 
\toprule
mMAPE (\%) & K-12$\downarrow$ & K-16$\downarrow$ & K-21$\downarrow$ & All$\downarrow$ \\ 
\midrule
OKS/PCK & 14.0 & 11.0 & 13.8 & 12.1 \\
PMP & \phantom{0}{\bfseries5.6} &  \phantom{0}{\bfseries5.4} & \phantom{0}{\bfseries6.0} &  \phantom{0}{\bfseries8.0} \\
\bottomrule
\end{tabularx}
\caption{Performance comparison between the OKS/PCK and PMP metrics for keypoint detection in phenotype measurement using mMAPE on common carp, representing the performance of the models selected by the different measures. ``All'' column represents the average mMAPE across all keypoints. The lower the mMAPE, the better the performance.}

\label{tab:effect_of_metric}
\end{table}



\begin{figure}[t!]
\centering
{\includegraphics[width=0.15\textwidth,height=0.11\textwidth]{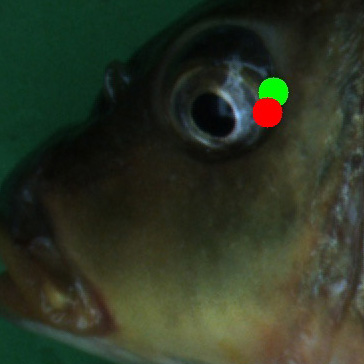}}
\hfill
{\includegraphics[width=0.15\textwidth,height=0.11\textwidth]{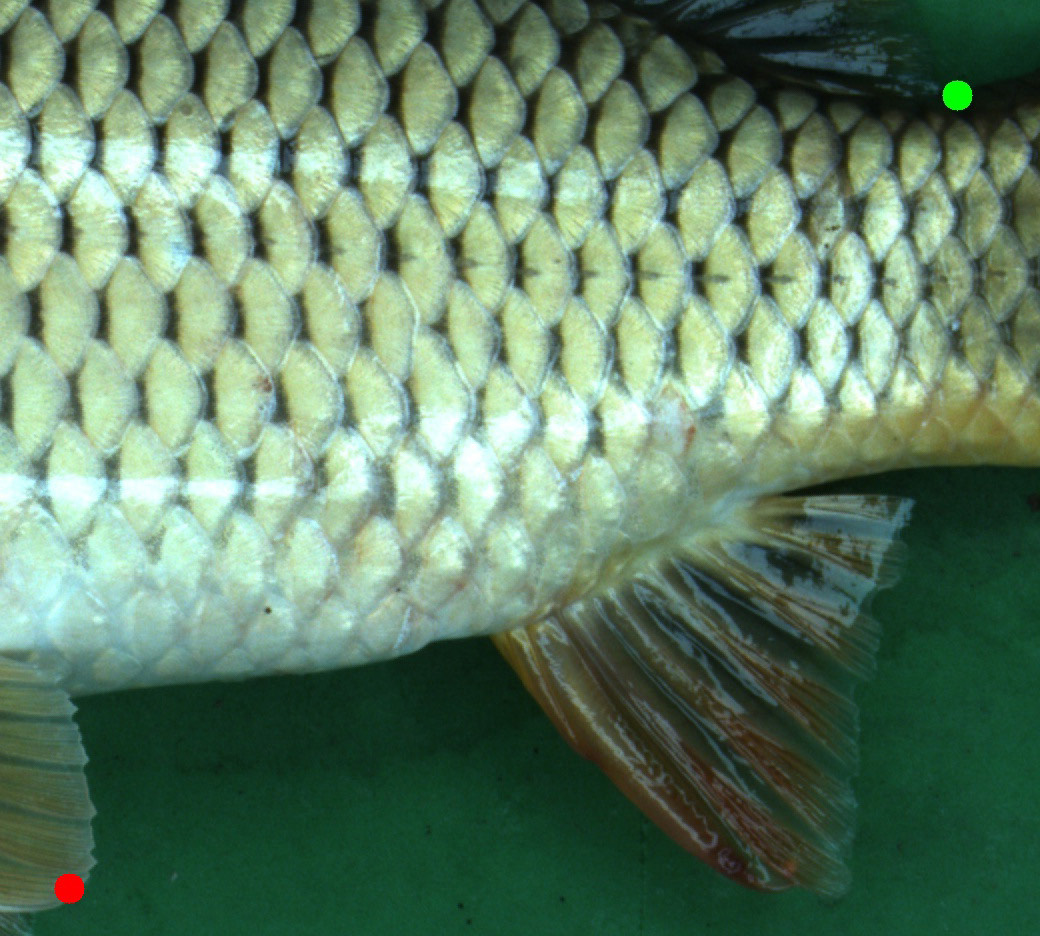}}
\hfill
{\includegraphics[width=0.15\textwidth,height=0.11\textwidth]{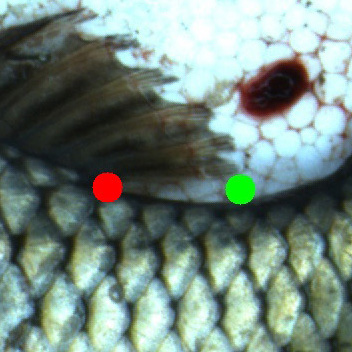}}

\subfloat[K-12]
{\includegraphics[width=0.15\textwidth,height=0.11\textwidth]{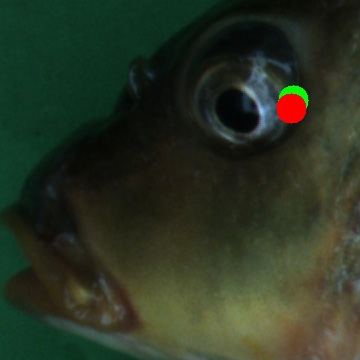}}
\hfill
\subfloat[K-16]
{\includegraphics[width=0.15\textwidth,height=0.11\textwidth]{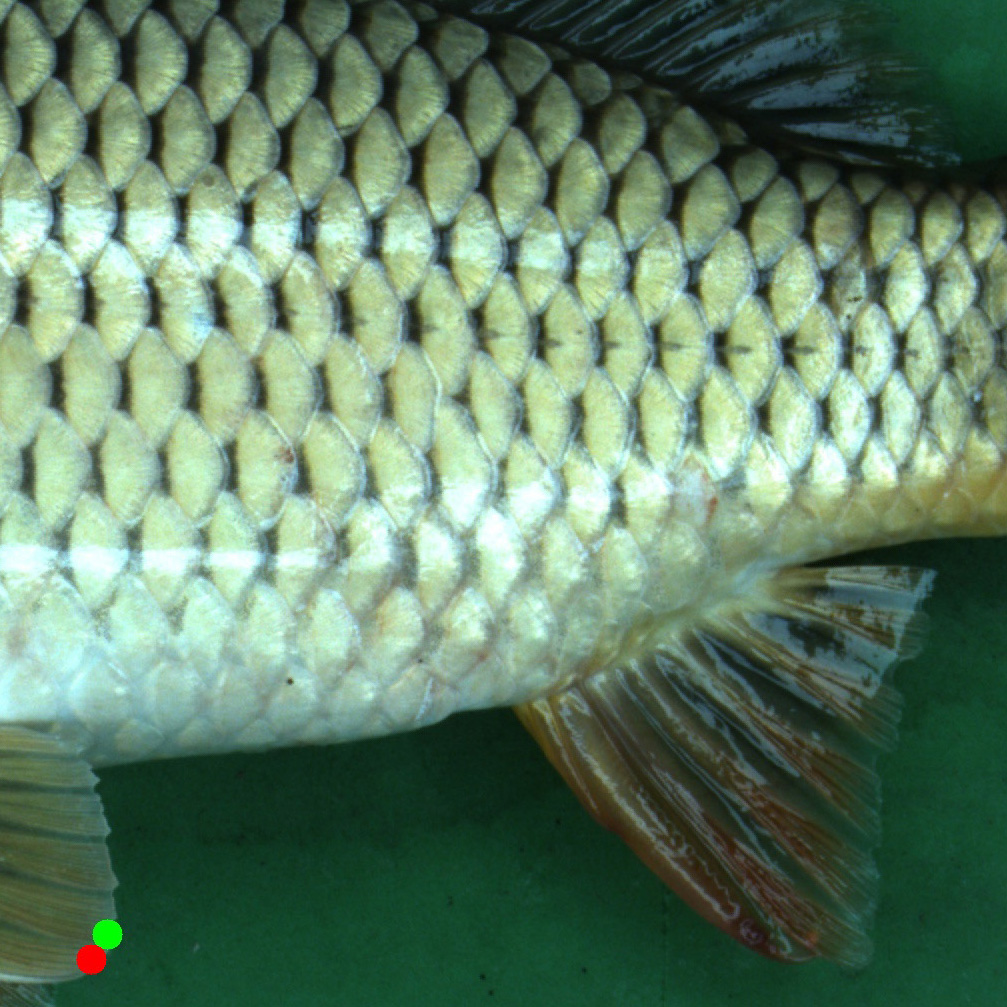}}
\hfill
\subfloat[K-21]
{\includegraphics[width=0.15\textwidth,height=0.11\textwidth]{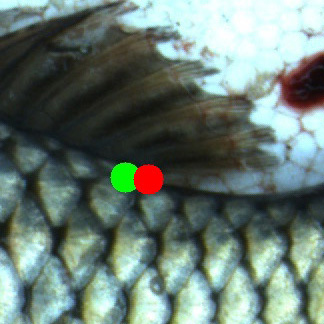}}
\caption{Visualizations of keypoint detection results using OKS/PCK and PMP as evaluation metrics on common carp. The first row shows the results of OKS/PCK and the second row shows the results of PCK. Three keypoints of interest are chosen due to their practical prediction difficulty. Under the same training protocol, the best model weights are selected under three evaluation metrics for prediction on the test images, where OKS and PCK select the same model. In each figure, the green dot is the predicted keypoint and the red dot is the annotated ground truth. PMP shows obvious better performance for these difficult keypoints.}
\label{fig:3x3subfig}
\end{figure}

\subsection{Effect of Percentage of Measured Phenotype}
We compare the performance of our proposed PMP against mainstream metrics OKS and PCK in keypoint detection for the fish phenotype measurement task. We train models using the common carp data and utilize these evaluation metrics to preserve the final models of keypoint detection. The Mean Absolute Percentage Error (MAPE) is employed as the indicator to evaluate the accuracy of phenotype measurements when using predicted keypoints of these selected models. We define the ground truth measurements of the phenotypes as those obtained based on the ground truth keypoints, and the predicted measurements of the phenotypes are derived from the model-predicted keypoints. It is worth noting that each keypoint may be used for measuring multiple phenotypes. Thereby, to assess the effectiveness of a keypoint in measuring all their derived phenotypes, we calculate its mean Mean Absolute Percentage Error (mMAPE), \ie~, averaging MAPEs for all associated phenotypes.

Tab.~\ref{tab:effect_of_metric} presents the mMAPE for three keypoints (K-12, K-16, and K-21) and its averaged value across overall 22 keypoints. We select these keypoints because they have a significant impact on our task. This is because the phenotypic measurements derived from these keypoints are of small size, making their accurate measurement the most challenging. In Tab.~\ref{tab:effect_of_metric}, we find that the keypoints predicted by the PMP-evaluated model have smaller mMAPE for both three keypoints and all keypoints, showing the superior evaluation capability of the PMP metric compared to the other two metrics. The identical values of mMAPE in OKS and PCK-based models are because both metrics utilize a uniform scaling factor when counting the correctly predicted keypoints.

\begin{table}[t] 
\centering
\captionsetup{font=footnotesize, width=\columnwidth, justification=justified}
\begin{tabularx}{\columnwidth}{@{}l *{4}{>{\centering\arraybackslash}X}@{}}
\toprule
PMP (\%) & K-6 $\uparrow$ & K-13 $\uparrow$ & K-17 $\uparrow$ & All $\uparrow$ \\
\midrule
Hourglass & 55.0 & 90.0 & 87.5 & 92.2 \\
Hourglass+ACR & \bfseries72.5 & \bfseries100.0\phantom{0} & \bfseries97.5 & \bfseries93.4 \\
\bottomrule
\end{tabularx}
\caption{Performance evaluation of ACR module incorporated in Hourglass network using the PMP metric on mottled naked carp. The ``All'' column indicates the average PMP across all keypoints. The higher the PMP, the better the performance.} 
\label{tab:effect_of_loss_hourglass}
\end{table}
\begin{table*}[t!]
\centering
\captionsetup{font=footnotesize, width=\textwidth, justification=justified}
\begin{tabularx}{\textwidth}{@{}l *{16}{>{\centering\arraybackslash}X}@{}}
\toprule
PMP (\%) & \multicolumn{4}{c}{Grouper} & \multicolumn{4}{c}{Mottled naked carp} & \multicolumn{4}{c}{Bighead  carp} & \multicolumn{4}{c}{Common carp} \\
\cmidrule(lr){2-5} \cmidrule(lr){6-9} \cmidrule(lr){10-13} \cmidrule(l){14-17}
& \scriptsize{K-6} & \scriptsize{K-11} & \scriptsize{K-12} & \scriptsize{All} & \scriptsize{K-6} & \scriptsize{K-15} & \scriptsize{K-18} & \scriptsize{All} & \scriptsize{K-5} & \scriptsize{K-8} & \scriptsize{K-11} & \scriptsize{All} & \scriptsize{K-6} & \scriptsize{K-11} & \scriptsize{K-19}&\scriptsize{All}
 \\
\midrule
HRNet & \footnotesize\phantom{0}95.0 & \footnotesize87.5 & \footnotesize85.0 &\footnotesize95.3 & \footnotesize\bfseries72.5 & \footnotesize\phantom{0}92.5 & \footnotesize87.5 & \footnotesize94.4 & \footnotesize57.5 & \footnotesize90.0 & \footnotesize77.5 & \footnotesize90.3 & \footnotesize52.5 & \footnotesize77.5 & \footnotesize85.0 & \footnotesize84.6 \\
HRNet+ACR & \footnotesize\bfseries100.0 & \footnotesize\bfseries92.5 & \footnotesize\bfseries90.0 & \footnotesize\bfseries95.6 & \footnotesize70.0 & \footnotesize\bfseries100.0 & \footnotesize\bfseries92.5 & \footnotesize\bfseries95.6 & \footnotesize\bfseries72.5 & \footnotesize\bfseries95.0 & \footnotesize\bfseries85.0 & \footnotesize\bfseries91.3 & \footnotesize\bfseries72.5 & \footnotesize\bfseries90.0 & \footnotesize\bfseries95.0 & \footnotesize\bfseries88.2 \\
\bottomrule
\end{tabularx}
\caption{Performance evaluation of ACR module incorporated in HRNet network using the PMP metric for four fish species. The ``All'' column provides the average PMP across all keypoints for each fish species. The higher the PMP, the better the performance.}
\label{tab:effect_of_loss_pmp}
\end{table*}
\begin{table*}[t!] 
\centering
\captionsetup{font=footnotesize, width=\textwidth, justification=justified}
\begin{tabularx}{\textwidth}{@{}l *{12}{>{\centering\arraybackslash}X}@{}}
\toprule
\multirow{2}{*}{Method} & \multicolumn{3}{c}{Grouper} & \multicolumn{3}{c}{Mottled naked carp} & \multicolumn{3}{c}{Bighead  carp} & \multicolumn{3}{c}{Common carp} \\
\cmidrule(lr){2-4} \cmidrule(lr){5-7} \cmidrule(lr){8-10} \cmidrule(r{0em}){11-13}
& \footnotesize MAPE$\downarrow$ (\%) &\footnotesize Corr$\uparrow$ (\%) &\footnotesize $R^2$$\uparrow$ (\%)
& \footnotesize MAPE$\downarrow$ (\%) &\footnotesize Corr$\uparrow$ (\%) &\footnotesize $R^2$$\uparrow$ (\%)
&\footnotesize MAPE$\downarrow$ (\%) &\footnotesize Corr$\uparrow$ (\%) &\footnotesize $R^2$$\uparrow$ (\%)
&\footnotesize MAPE$\downarrow$ (\%) &\footnotesize Corr$\uparrow$ (\%) &\footnotesize $R^2$$\uparrow$ (\%)\\
\midrule
HRNet & \footnotesize2.8 & \footnotesize86.2 & \footnotesize76.6 & \footnotesize3.7 & \footnotesize72.5 & \footnotesize61.2 & \footnotesize9.9 & \footnotesize77.0 & \footnotesize70.0 & \footnotesize6.6 & \footnotesize80.9 & \footnotesize68.7 \\
HRNet+ACR & \footnotesize\bfseries2.7 & \footnotesize\bfseries89.0 &\footnotesize\bfseries80.6 & \footnotesize\bfseries2.6 & \footnotesize\bfseries90.8 & \footnotesize\bfseries83.8 & \footnotesize\bfseries7.4 &\footnotesize\bfseries82.3 & \footnotesize\bfseries75.7 & \footnotesize\bfseries4.7 & \footnotesize\bfseries87.3 & \footnotesize\bfseries78.6 \\
\bottomrule
\end{tabularx}
\caption{Performance evaluation of ACR module for four fish species using phenotypic performance metrics. The metrics included MAPE, ``Corr'' for Pearson correlation coefficient, and $R^2$. Lower values of MAPE indicate better performance, whereas higher values of Pearson correlation coefficient and $R^2$ indicate better performance. }
\label{tab:effect_of_loss_phenotypes}
\end{table*}

Fig.~\ref{fig:3x3subfig} provides a qualitative analysis that further reveals this finding. A visual comparison across the three keypoints highlights the superiority of the PMP metric over OKS and PCK in terms of keypoint detection accuracy, where the prediction is located closer to the annotated ground truth. 

\subsection{Effect of ACR for Keypoint Detection}

\noindent \textbf{Sensitivity to Network Backbone.}
To evaluate the effectiveness of ACR for keypoint detection, we first explore its robustness under different network architectures. Two representative models are chosen as our baselines, namely: Hourglass~\cite{newell2016stacked} and HRNet~\cite{wang2020deep}. The performance of each individual keypoint, as well as the overall performance across all keypoints, is evaluated using the PMP metric.

By examining the results presented in Tab.~\ref{tab:effect_of_loss_hourglass} and Tab.~\ref{tab:effect_of_loss_pmp}, we can observe the comparative outcomes of the Hourglass and HRNet models with and without ACR, respectively. It becomes evident that incorporating the ACR module enhances the performance of both networks, proving its remarkable ability to generalize effectively.

\noindent \textbf{Robustness on Species}.
We explore the robustness of ACR across different fish species. For HRNet, we conducted comparative analyses with and without the ACR module on four fish species in our FishPhenoKey dataset. The results in Tab.~\ref{tab:effect_of_loss_pmp} clearly demonstrate that the incorporation of the ACR module improves the overall performance for all keypoints for the four fish species, as evidenced by the increased PMP values. This finding confirms the effectiveness and robustness of ACR across different species.

\subsection{Effect of ACR for Morphological Assessment}
This section is dedicated to evaluating the effectiveness of the ACR module in terms of morphological phenotype measurements. We use the MAPE and the Pearson correlation coefficient for assessing the accuracy of phenotype measurements by comparing ground truth measurements with predicted measurements. The models used for keypoint detection are selected based on the PMP metric. Additionally, a simple linear regression is performed for each phenotype in each fish species, fitting the ground truth and predicted measurements. The determination coefficient ($R^2$) from the regression, widely adopted in bioimage analysis studies~\cite{zhu2021deep,tang2023strategy}, serves as an indicator to assess the consistency between the ground truth and predicted measurements. These metrics enable us to validate the effectiveness of our loss module and quantify the improvements over the baseline, providing precise enhancement metrics for phenotypic predictions. 

As depicted in Tab.~\ref{tab:effect_of_loss_phenotypes}, the incorporation of ACR into the HRNet framework markedly enhances the performance of phenotype measurements for all four fish species. This is evidenced by reduced MAPE values and increased correlation and $R^2$ values of HRNet+ACR model compared to HRNet, indicating higher accuracy, correlation, and consistency between the the ground truth measurements and predicted measurements. The data presented in Tab.~\ref{tab:effect_of_loss_phenotypes} are the average values across all 23 phenotypes, indicating that HRNet+ACR outperforms HRNet when considering the overall performance across all phenotypes.

\begin{figure}[t]
\centering
\subfloat[HRNet]{\includegraphics[width=0.47\columnwidth]{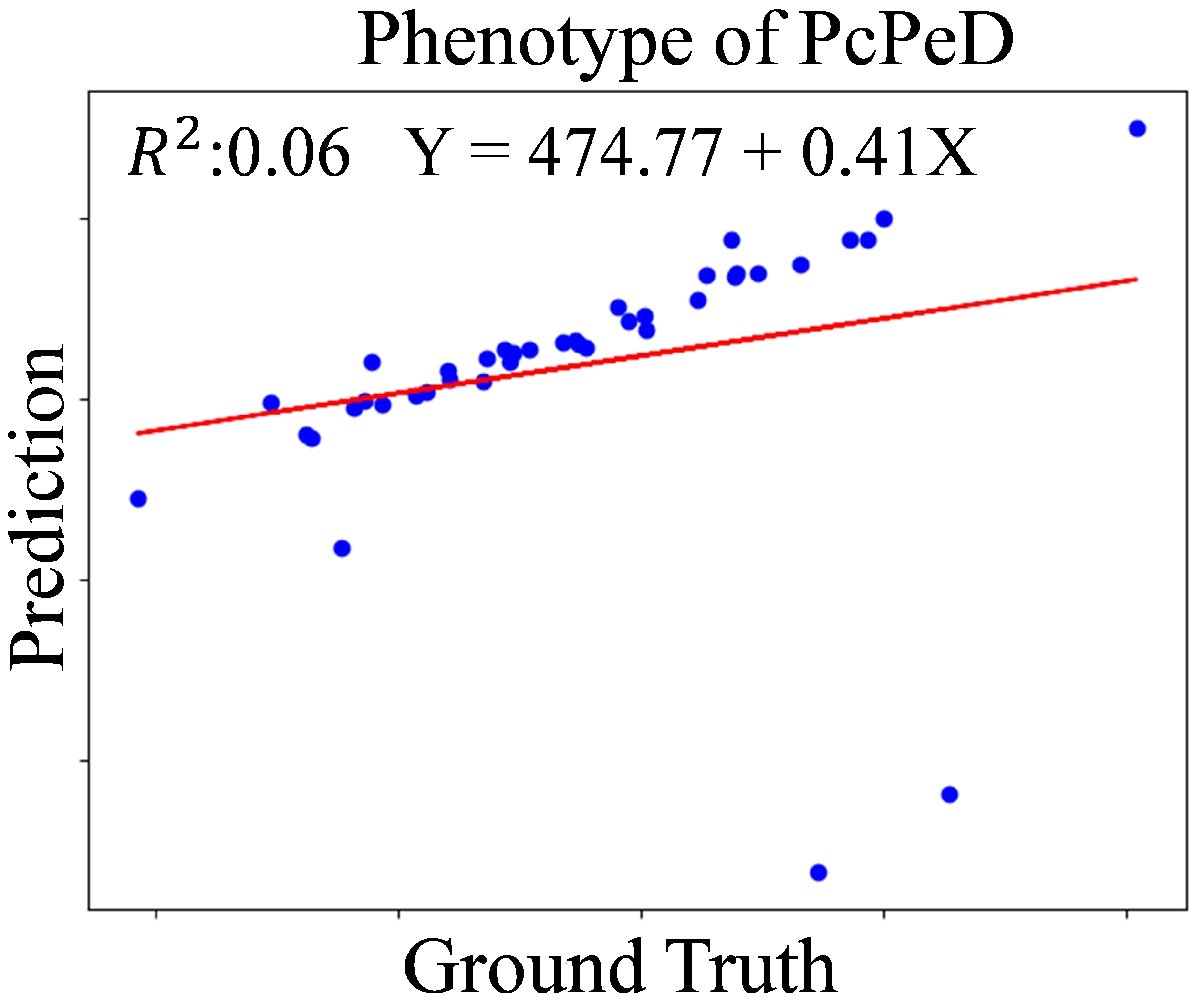}}\hfill
\subfloat[HRNet+ACR]{\includegraphics[width=0.47\columnwidth]{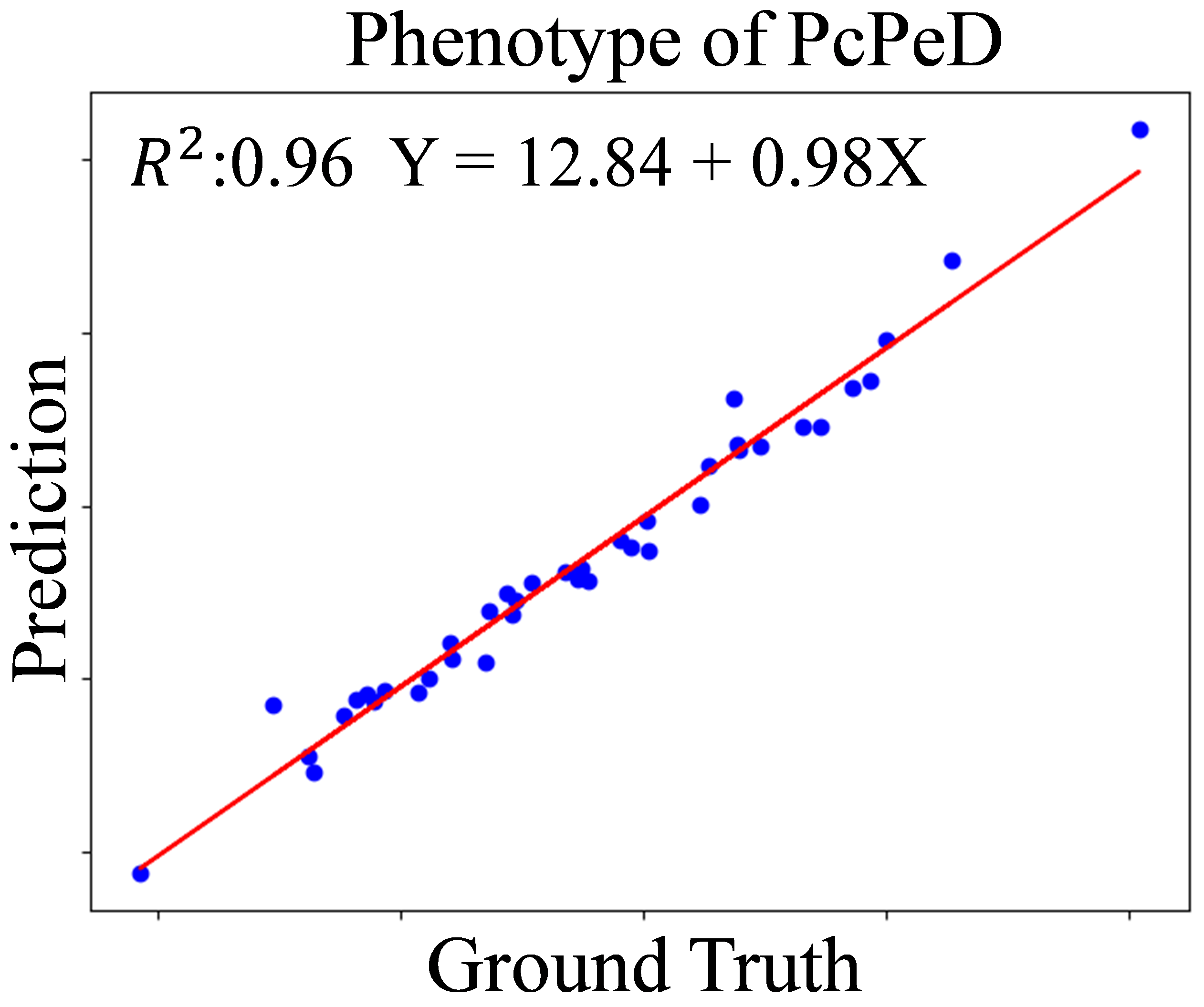}}
\caption{Consistency comparisons between ground truth measurement and predicted measurements of the phenotype of PcPeD using HRNet and HRNet+ACR for mottled naked carp. In scatterplots, each blue dot represents a measurement of each fish, which is used to fit the simple linear regression (represened by the red line) with its equation and determination coefficient $R^2$ provided.}
\label{fig:R2}
\end{figure}

Upon closer examination of individual phenotypes, we analyze scatterplots depicting the consistency between ground truth measurements and predicted measurements using both the HRNet and HRNet+ACR models, as shown in Fig.~\ref{fig:R2}. In the HRNet plot, several noticeable deviations can be observed, with a low $R^2$ of 6\%. However, in contrast, the HRNet+ACR plot exhibits good consistency with a high $R^2$ value of 96\%. This trend suggests that the integration of the ACR module enhances the robustness of the algorithm when measuring the same phenotype across different fish images.

\section{Conclusion}
In this work, we introduce a new keypoint detection method for the accurate measurement of fish phenotypes. We provide the meticulously annotated FishPhenoKey dataset, develop the PMP metric that accurately measures phenotypes according to the biological and phenotypic scale of keypoints, and integrate the ACR module that enhances detection precision. These contributions are pivotal for advancing precise fish phenotype measurement and have significant implications for sustainable aquaculture and biodiversity research.




\clearpage

\clearpage
\section*{Acknowledgements}
This study was partially supported by the National Science Foundation of China (No.32200331, No.32090061), the Major Science and Technology Research Project of Hubei Province (No.2021AFB002), the Start-Up Grant from Wuhan University of Technology of China (No.104-40120526), and Shanghai Artificial Intelligence Laboratory.
\bibliographystyle{named}
\bibliography{refs}

\newpage
\appendix

\section{Source Code}
The source code is provided in \texttt{\url{https://github.com/WeizhenLiuBioinform/Fish-Phenotype-Detect}} to re-implement the reported results, including detailed documentation of the data processing and training protocol.

\section{Definition of Phenotypes}
Tab.~\ref{tab:defpheno} lists 23 distinct phenotypes, each defined by specific keypoint pairs among 22 key locations, with the first column presenting the abbreviation of each phenotype, the second detailing the full phenotype name, and the third indicating the keypoint pairs used for their calculation.The keypoint pairs are as follows: 
\begin{table}[ht]
\footnotesize
\begin{tabular}{|l|l|l|}
\hline
\scriptsize\textbf{Abbreviation} & \scriptsize\textbf{Phenotypes} & \scriptsize\textbf{Keypoints} \\
\hline
TL & total length & K-1, K-9 \\
SL & standard length & K-1, K-10 \\
HL & head length & K-1, K-2 \\
SnL & snout length & K-1, K-11 \\
ED & eye diameter & K-11, K-12 \\
PoL & postorbital length & K-12, K-2 \\
BD & body depth & K-5, K-6 \\
HD & head depth & K-3, K-4 \\
PeAD & pelvic–anal fin origin distance & K-15, K-17 \\
CPD & caudal peduncle depth & K-7, K-8 \\
CPL & caudal peduncle length & K-18, K-10 \\
DFL & dorsal fin length & K-20, K-21 \\
DFH & dorsal fin height & K-20,  K-22 \\
PcL & pectoral fin length & K-13, K-14 \\
PeL & pelvic fin length & K-15, K-16 \\
AFL & anal fin length & K-17, K-18 \\
AFH & anal fin height & K-17, K-19 \\
TFL & tail fin length & K-10, K-9 \\
PrDL & predorsal length & K-1, K-20 \\
PoDL & postdorsal length & K-20, K-10 \\
PcDD & pectoral–dorsal fin origin distance & K-13, K-20 \\
PcPeD & pectoral–pelvic fin origin distance & K-13, K-15 \\
PeDD & pelvic–dorsal fin origin distance & K-15, K-20 \\
\hline
\end{tabular}
\caption{ Comprehensive overview of phenotypes and their corresponding keypoints.}
\label{tab:defpheno}
\end{table}

\begin{figure}[t!]
 \centering
 \begin{subfigure}[b]{\columnwidth}
   \centering
   \includegraphics[width=\textwidth,height=2.5cm]{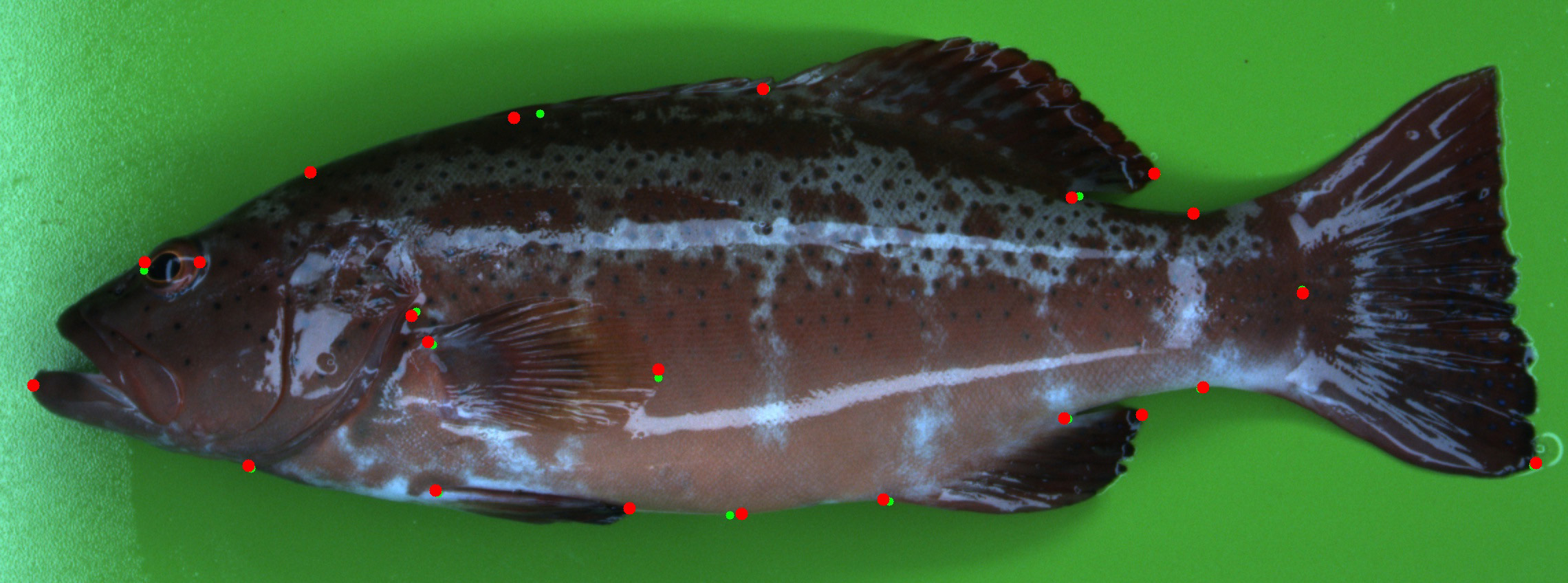}
   \caption{Grouper}
   \label{fig:grouper}
 \end{subfigure}
 
 \begin{subfigure}[b]{\columnwidth}
   \centering
   \includegraphics[width=\textwidth]{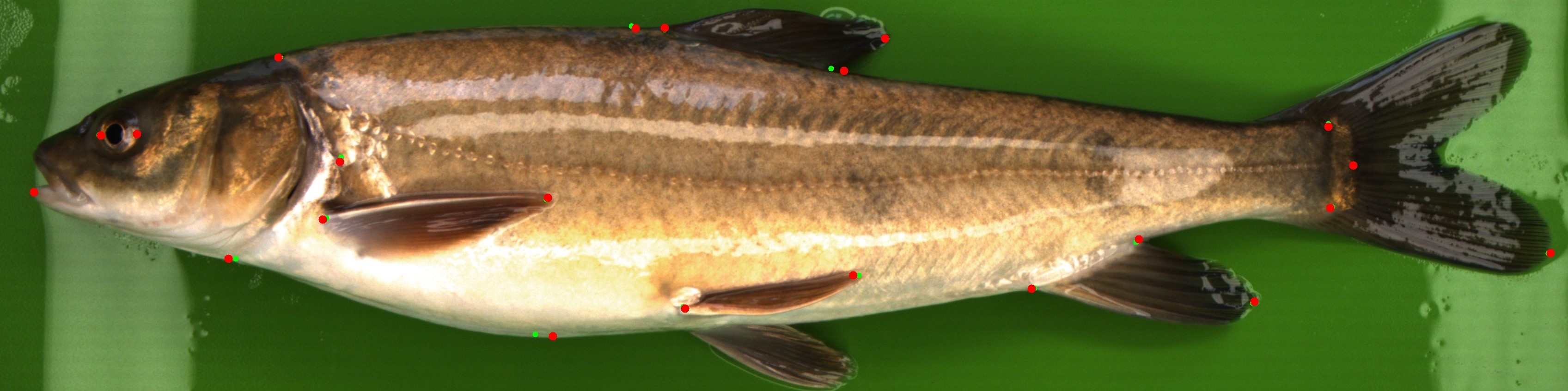}
   \caption{Mottled naked carp}
   \label{fig:mottled_naked_carp}
 \end{subfigure}
 
 \begin{subfigure}[b]{\columnwidth}
   \centering
   \includegraphics[width=\textwidth]{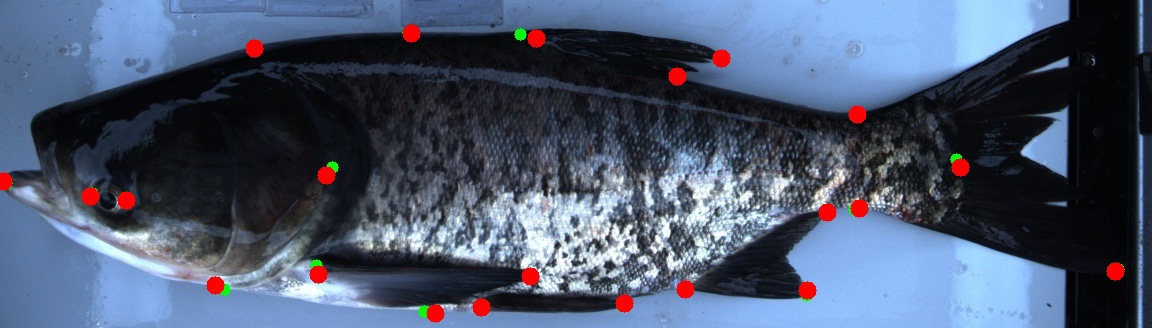}
   \caption{Bighead carp}
   \label{fig:bighead_carp}
 \end{subfigure}
 
 \begin{subfigure}[b]{\columnwidth}
   \centering
   \includegraphics[width=\textwidth]{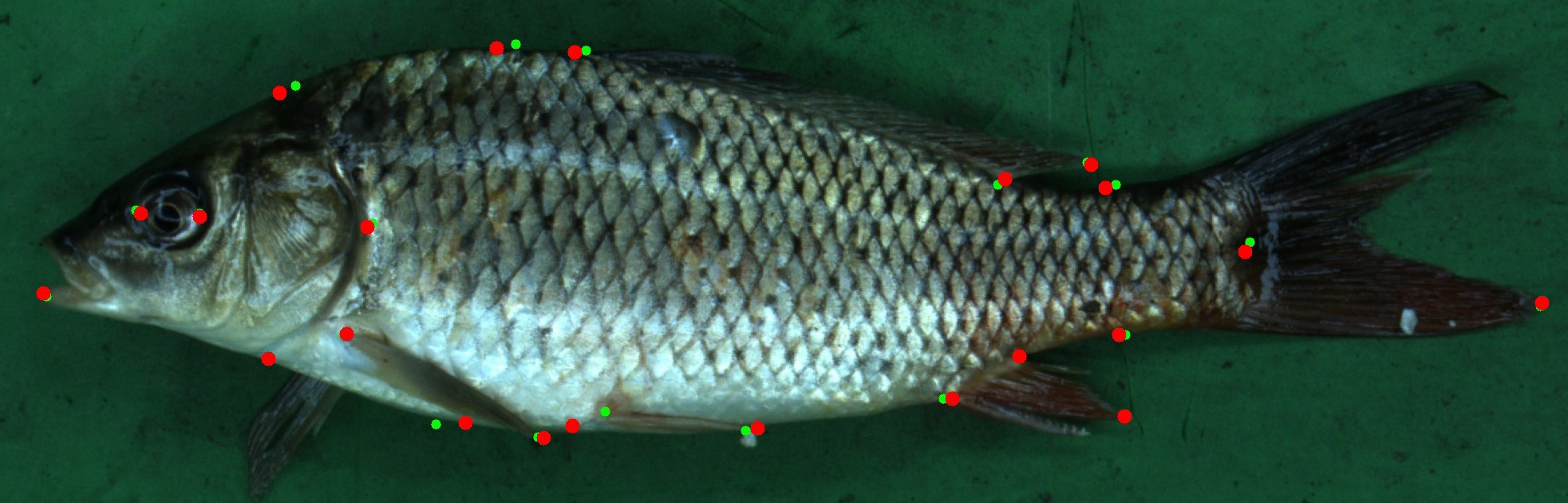}
   \caption{Common carp}
   \label{fig:common_carp}
 \end{subfigure}
 
 \caption{Comparative visualization of predicted and actual keypoints for four fish species, with each image showcasing an overlay of predicted (green) and actual (red) points.}
 \label{fig:fish_keypoints}
\end{figure}
\section{Additional Experimental Details}

This study implements three data augmentation techniques: random scaling, random rotation, and random translation. The images of bighead carp are resized to dimensions of $1152\times864$ pixels, while the images of mottled naked carp, common carp, and grouper are resized to a larger resolution of $4608\times3456$ pixels. The augmentations include random rotation at an angle of 45\textdegree\ and random scaling within a range of plus or minus 35\%.

\section{Additional Visualization Analysis}
In fig.~\ref{fig:fish_keypoints}, we analyze the visualization results of our Anatomically Calibrated Regularization (ACR) method for four different fish species. The experimental design is consistent with Section 6 in the main text. We find that the predicted values are very close to the actual values, thereby meeting the requirements of the phenotypic measurement task.

\end{document}